\documentclass[10pt,twocolumn,letterpaper]{article}

\usepackage[pagenumbers]{cvpr} 

\usepackage[utf8]{inputenc} 
\usepackage[T1]{fontenc}    
\usepackage{xr-hyper}   
\usepackage{url}            
\usepackage{booktabs}       
\usepackage{amsfonts}       
\usepackage{nicefrac}       
\usepackage{microtype}      
\usepackage{xcolor}         

\usepackage{lipsum} 

\usepackage{times}
\usepackage{latexsym}
\usepackage{enumitem}
\setlist[itemize]{itemsep=0.5ex, topsep=0.3ex, parsep=0pt, partopsep=0pt}
\makeatletter
\renewcommand{\section}{%
  \@startsection{section}{1}{\z@}%
    {0.8ex plus 0.4ex minus .1ex}
    {0.8ex}
    {\normalfont\large\bfseries}%
}

\renewcommand{\subsection}{%
  \@startsection{subsection}{2}{\z@}%
    {0.6ex plus 0.3ex minus .1ex}
    {0.5ex}
    {\normalfont\normalsize\bfseries}%
}
\makeatother

\usepackage{natbib}

\usepackage{amsmath}
\usepackage{amsfonts}
\usepackage[ruled,vlined]{algorithm2e}
\usepackage{booktabs}
\usepackage{multirow, colortbl}

\usepackage{tikz}
\usepackage{colortbl}
\usepackage{makecell}
\usepackage{tabularx}
\usepackage{caption}
\usepackage{graphicx}
\usepackage{caption}

\usepackage{tikz}
\usetikzlibrary{positioning,calc,decorations.pathreplacing}

\usepackage[most]{tcolorbox}

\usepackage{smartdiagram}
\usepackage{amssymb}  
\usepackage{graphicx} 
\usepackage{lipsum}

\definecolor{cvprblue}{rgb}{0.21,0.49,0.74}
\usepackage[pagebackref,breaklinks,colorlinks,allcolors=cvprblue]{hyperref}

\def\paperID{} 
\def\confName{CVPR}
\def\confYear{2026}

\title{Align While Search: Belief-Guided Exploratory Inference \\for World-Grounded Embodied Agents}

\author{
Seohui Bae\textsuperscript{1} \quad
Jeonghye Kim\textsuperscript{2} \quad
Youngchul Sung\textsuperscript{2} \quad
Woohyung Lim\textsuperscript{1}\\[0.3em]
\textsuperscript{1}LG AI Research \quad
\textsuperscript{2}KAIST\\[0.3em]
{\tt\small \{seohui.bae,w.lim\}@lgresearch.ai},
{\tt\small \{jeonghye.kim,ycsung\}@kaist.ac.kr}\\
}

\begin{document}
\maketitle

\begin{abstract}
In this paper, we propose a test-time adaptive agent that performs exploratory inference through posterior-guided belief refinement without relying on gradient-based updates or additional training for LLM agent operating under partial observability.  
Our agent maintains an external structured belief over the environment state, iteratively updates it via action-conditioned observations, and selects actions by maximizing predicted information gain over the belief space.
We estimate information gain using a lightweight LLM-based surrogate and assess world alignment through a novel reward that quantifies the consistency between posterior belief and ground-truth environment configuration.
Experiments show that our method outperforms inference-time scaling baselines such as prompt-augmented or retrieval-enhanced LLMs, in aligning with latent world states with significantly lower integration overhead. 
\end{abstract}  

\section{Introduction} 

Agents operating in partially-observable environments continuously encounter incomplete information in the course of accomplishing their goals~\citep{kaelbling1998planning,igl2018deep}. The key capability required in such settings is exploratory decision making: The agent should not only act to achieve the objective but also to collect information that refines its  belief about the world. The interaction between action, observation, and belief refinement forms the basis of effective behavior under uncertainty.

\begin{figure}[t]
    \centering    \includegraphics[width=\linewidth]{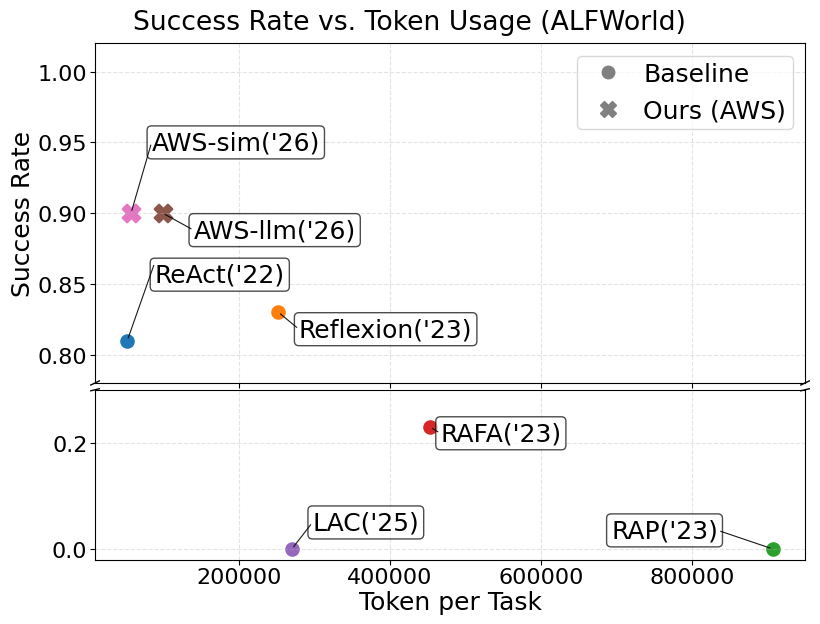}
    \caption{\textbf{Success rate vs. token usage on ALFWorld subtasks}. AWS achieves higher success with 2-5$\times$ fewer tokens compared to strong inference-time baselines, highlighting efficient belief-guided exploration under partial observability. Full results in Appendix \textcolor{blue}{20}.}
    \label{fig:performance}
    \vspace{-20pt}
\end{figure} 

In this paper, we consider such exploratory decision making under uncertainty with focus on search task, which is the basic task for more advanced tasks, and examine inference-time world understanding under partial observability.
While large language models (LLMs) have shown promise in zero-shot task execution~\citep{brown2020language, openai2023gpt4}, their static reasoning often fails to adapt to the unfolding dynamics of environments with partial observability. Prior approaches to such partial observability include 
train-time policy optimization via supervised or reinforcement learning training~\citep{ouyang2022training, rafailov2023direct, yao2022react, gao2022pal} and inference-time scaling methods~\citep{li2021prefix, lewis2020retrieval, ram2023context}.
However, the former requires substantial training costs and limits deployment flexibility, while the latter lacks adaptive interaction with the environment.
More recent inference-time agents, such as Reflexion~\citep{shinn2023reflexion}, RAP~\citep{hao2023reasoning}, RAFA~\citep{liu2024reason}, LAC~\citep{dong2025enhancing}, and ReflAct~\citep{kim2025reflact}, couple LLMs with classical MDP planning or question-based belief summaries, but typically rely on additional simulators or learned critics and do not maintain an explicit probability distribution over latent environment configurations and object locations.  We instead view exploratory decision making for search under partial observability as an \emph{approximate Bayes-adaptive control} problem: the agent maintains a posterior over latent environment configurations and selects actions that trade off task reward and information gain in belief space.

With this  perspective,  we propose a novel, lightweight but effective agent architecture that performs \emph{inference-time exploratory reasoning} through {\em posterior-guided belief refinement}. Our core idea is as follows: the agent maintains a structured posterior belief over search action space, updates it based on environmental feedback, and selects actions that reflect this evolving belief. Thus, our behavior policy is driven not by memorized patterns but by a dynamically refined understanding of the world. 
Crucially, all our adaptation occurs at test time without gradient updates, fine-tuning, or additional environment models. 
Empirical results show that our inference-time alignment strategy outperforms inference-time policy optimization baselines with substantially lower computational overhead, as shown in Fig~\ref{fig:performance}. 
Our method generalizes across diverse object types, environments, and interaction histories without requiring task-specific training or reward-based tuning. 

Our contributions are summarized below:

\noindent $\bullet$ We introduce \emph{Align While Search} (AWS), a lightweight world-aligned agent that casts test-time control as belief-guided  search over object-centric hypotheses, without any additional training or gradient updates.

\noindent  $\bullet$ We provide a Bayesian account of AWS by viewing it as approximate Bayes-adaptive control: a latent generative model with amortized posterior belief refinement implemented by an LLM, and information-gain-driven action selection on the belief state.

\noindent  $\bullet$ Across ALFWorld, VirtualHome, and BabyAI, in both text-only and image-augmented settings, we show that AWS consistently improves search success–cost trade-offs over inference-time prompt/RAG scaling and train-time world-model or policy-gradient baselines, as seen in Fig.~\ref{fig:performance}.

\section{Preliminaries}~\label{sec:preliminary}
\vspace{-25pt}
\paragraph{MDP/POMDP for LLM agents.}
Many language-agent settings can be cast as a Markov Decision Process (MDP) $\langle\mathcal{S}, \mathcal{A}, T, \mathcal{R}, \gamma \rangle$. In embodied environments such as ALFWorld and VirtualHome, the (latent) world state $s$ includes agent pose, object locations, container open/close states, and inventory. The agent issues high-level action $a$ (\textit{e.g.}, navigation, opening, manipulation) and receives textual observations $o$ describing the partially observed scene. 
Then, the world state $s$ changes according to transition $T$ based on the current state and agent action.  
For an LLM-based agent, the input is typically the interaction history~\citep{yao2022react}, $h_t=\{(o_i,a_i)\}_{i=1}^t$, or a textual/symbolic summary~\citep{shinn2023reflexion}, and the LLM implements a policy $\pi(a_t\mid h_t)$. Concrete instantiations for ALFWorld/VirtualHome are given in Appendix \textcolor{blue}{9.2}.
\vspace{-10pt}
\paragraph{Epistemic uncertainty and Bayes-adaptive MDPs.}
Because the environment is only partially observed, LLM agents in ALFWorld and VirtualHome are naturally modeled as POMDPs~\citep{kaelbling1998planning}: at each step the agent receives partial observation $o \sim \mathcal{O}(o\mid s,a)$ about the environment state $s$. Beyond perceptual uncertainty, there is also \emph{epistemic} uncertainty about latent environment parameters, such as object layouts or stochastic dynamics, which must be inferred from interaction. Bayes-adaptive MDPs (BAMDPs)~\citep{zhang2025beyond, jeong2025reflect} make this explicit by introducing latent parameters $\phi \in \Phi$ with prior $b_0(\phi)$ and treating the belief over $\phi$ as part of the state. Thus, in a BAMDP $\langle \mathcal{S}, \mathcal{A}, \Phi, T, \mathcal{R}, \gamma, b_0 \rangle$, the agent seeks a policy  that maximizes expected return with posterior updates over $\phi$ induced by the history, thereby motivating information-seeking behavior.

\section{Pitfalls and Opportunities in LLM Agents} \label{sec:motivation} 

\noindent Modern language agents, both   base models and those post-trained with supervised trajectories, tend to overfit to trained  behaviors. In this section, we investigate this phenomenon in partially observed settings.

\subsection{Failure modes in search tasks}\label{sec:pitfalls}
\begin{figure}[htbp]
\vspace{-12pt}
    \centering
    \includegraphics[width=0.9\columnwidth]{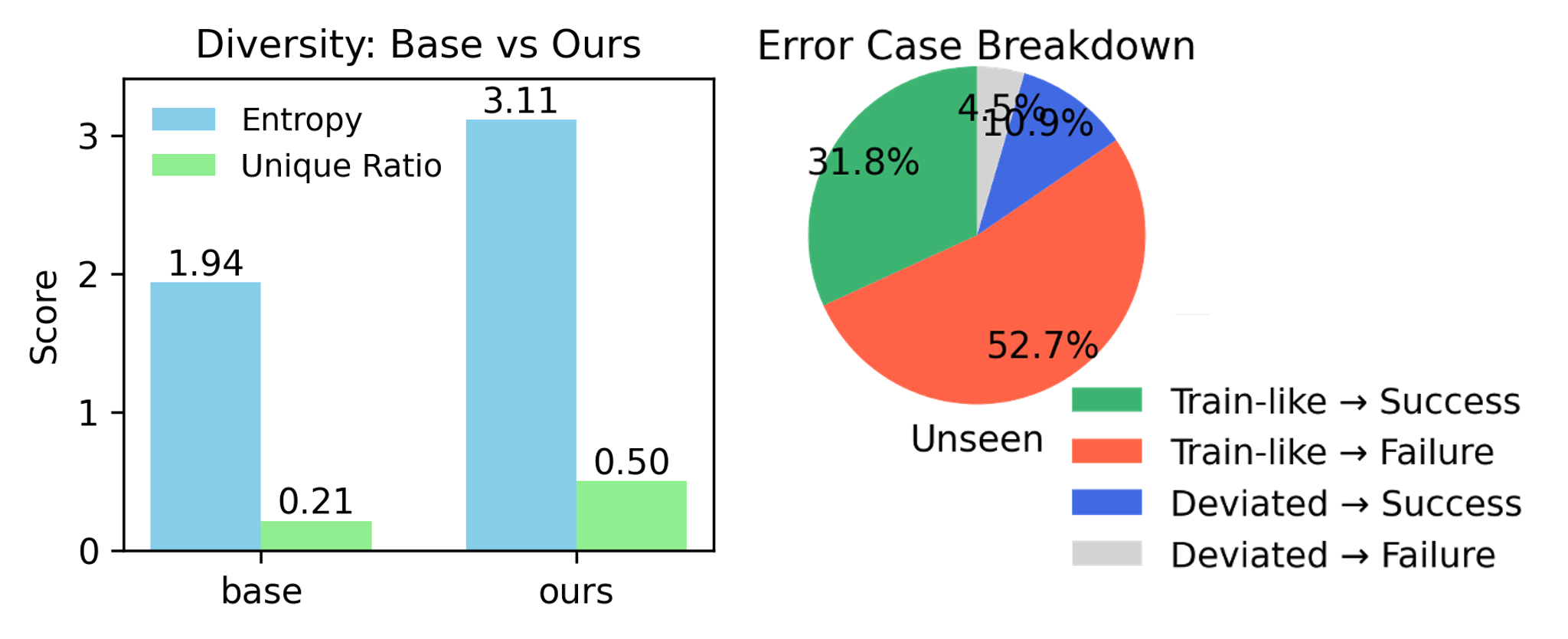}
    \vspace{-7pt}
    \caption{
    \textbf{Exploration Failure of Base and SFT Models.} 
    (\textbf{Left}) Trajectory diversity measured by entropy and the distinct trajectory ratio. 
    (\textbf{Right}) A breakdown of SFT agent failures reveals that most errors occur due to rigid replay of train-time search patterns.
    }
    \vspace{-12pt}
    \label{fig:mini-error-breakdown}
    
\end{figure}

\noindent We conducted an experiment to investigate the diversity and test-time adaptation of generated action sequences of an agent. In each run of this experiment, the agent needs to search for an object in a household, yielding one action sequence, and each run has a different room layout and a target location. The result is shown in Figure~\ref{fig:mini-error-breakdown} (left), where the entropy represents the entropy of empirical distribution on the set of actions computed from all experiment runs, and the unique ratio represents the number of distinct action sequences out of total action sequences.
It is seen that  the base model (GPT-4o-mini) exhibits low action entropy (1.94) and low distinct trajectory ratio (0.21) compared to our method 3.11 and 0.5, respectively, revealing that the base LLM agent yields repetitive search behavior with low diversity in action, even though the environment changes run by run. More importantly, we observed that this behavior persists even after supervised fine-tuning (SFT) by which the base model is further trained on expert trajectories.

Fig. ~\ref{fig:mini-error-breakdown} (right) and Appendix \textcolor{blue}{10} focus on the case that the agent is given search tasks in unseen test-time environments in which the room layout is similar to the trained layout but object locations are different. In this case, the agent persistently reproduces train-time room visitation sequences  in the test-time environments with 84.5\% (=52.7+31.8) out of total search sequences. Notably, over 50\% of failures occur in the case of train-like search sequences, indicating that the agent is not adapting based on what it observes but performs blindly as it is trained.

\vspace{-1pt}
\subsection{Opportunities for belief-augmented search}\label{sec:opportunities}
\noindent Introducing and inferring a latent variable $\phi$ representing the environment type or configuration can enhance  adaptation and efficient decision-making. It allows the agent to select context-appropriate actions, reduce uncertainty through targeted exploration, and generalize across similar environments. With the latent variable $\phi$, given a trajectory $\tau_t = (o_0, a_0, o_1, \dots, a_t)$, the agent maintains an explicit belief $b:=p(\phi|\tau)$ and selects actions to maximize expected information gain on the belief~\citep{lindley1956measure, degroot2005optimal, chaloner1995bayesian}.  Our key observation in this paper  is that many task environments have well-defined
 latent semantic structure, such as rooms differing by type (e.g., kitchen vs. bedroom) and associated object patterns so that we can adopt latent variable $\phi$ for such LLM agent's environments. 
\begin{figure}[htbp]
\vspace{-10pt}
    \centering
    \includegraphics[width=0.9\columnwidth]{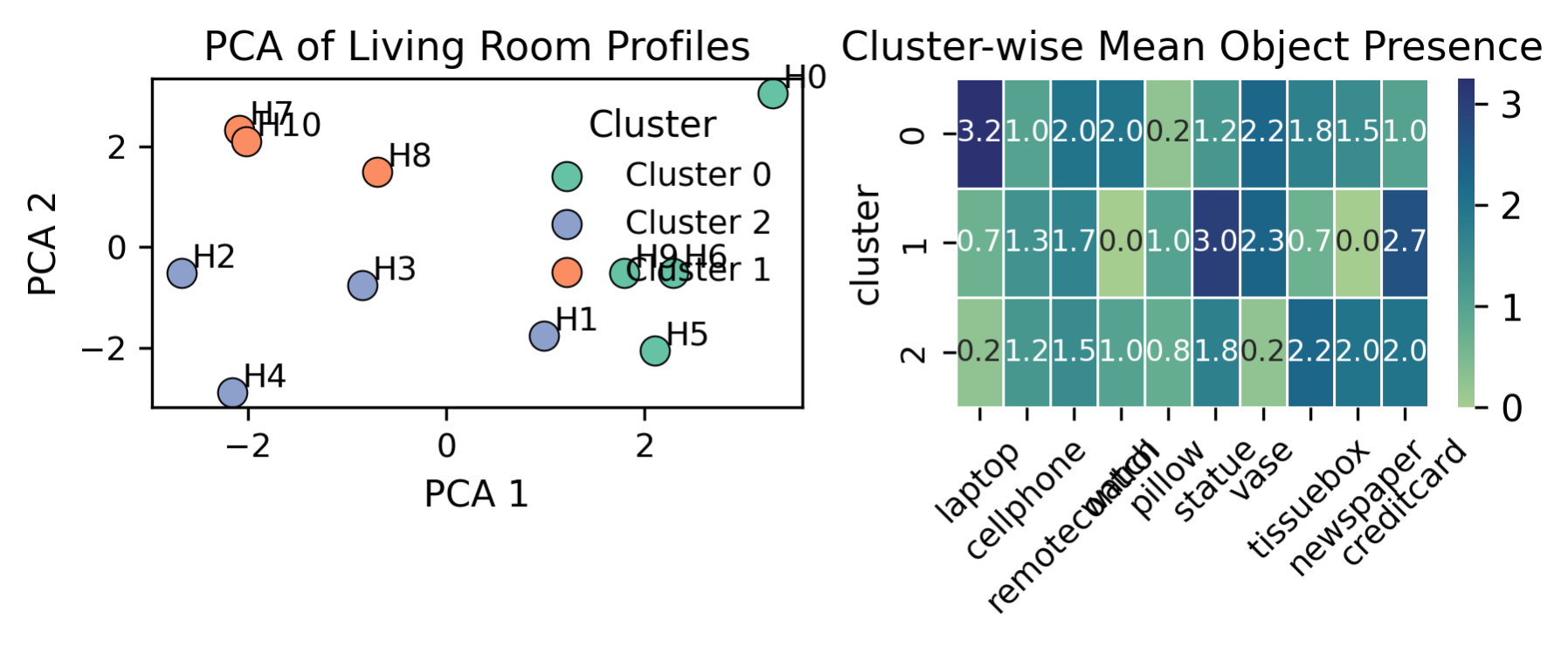}
    \vspace{-15pt}
    \caption{
        \textbf{Object Counts per House Reveal Latent Diversity.}
(\textbf{Left}) PCA shows clustered object usage patterns across houses.  
(\textbf{Right}) Each cluster exhibits distinct object preferences, indicating latent user behavior differences.
}
    \label{fig:heatmap_causal}
    \vspace{-12pt}
\end{figure}
 
 For example, as shown in Fig. \ref{fig:heatmap_causal} (left), households form natural clusters based on their living room object usage, while their corresponding cluster-wise object profile (Fig.~\ref{fig:heatmap_causal} (right)) reveals structured differences (e.g., tech-heavy or minimalist usage).  
  The observations in Fig. \ref{fig:heatmap_causal} suggest that, during search, the difficulty does not stem from complex world configuration or dynamics but from the agent's inability to efficiently exploit this well-defined latent structure. By adopting  the latent semantic variable  $\phi$ and constructing an explicit belief structure on $\phi$, we can  facilitate search task.  In the next section, we formalize this idea.

\section{Problem Formulation: Belief-Augmented Search}

\noindent We now introduce a formal abstraction of search task, which factors out world dynamics and focuses on the agent's epistemic uncertainty about object locations.

\vspace{-1em}
\paragraph{Single-State MDP for Search in Embodied Worlds}\label{sec:single_state_mdp}

Search tasks occurs repeatedly inside ALFWorld and VirtualHome episodes: the LLM agent must locate a target object before it performs any actions that change the world (\textit{e.g.,} moving the objects, rearranging containers).
At the search stage,  the physical world state (including all object locations) is static; what changes over time is  the agent's \emph{belief} about where the target is located.  Rather than modeling full embodied dynamics during this phase, we follow recent object-search formulations that abstract away world dynamics and operate purely over a belief on candidate locations~\citep{wandzel2019multi,kurenkov2023modeling}. Thus, we model the search task  as a belief-augmented single-state decision problem~\citep{stone1976theory, wandzel2019multi,zheng2022towards}. Formally, fix a search subgoal (\textit{e.g.,} find the apple) within an episode, and let $\mathcal{L}=\ell_1, \dots, \ell_L$ denote the finite set of candidate locations or receptacles where the target might reside (rooms, containers, or surfaces determined by the underlying environment). We introduce an explicit belief state $b_t \in \Delta^{L}$, where $b_t(\ell)$ is the agent's probability that the target object is at location $\ell$ at search step $t$ and $\Delta^L$ is the set of all probability distributions over $\mathcal{L}$.

During a search task, we abstract the underlying world state $s$ as fixed, and define a single-state MDP \[
\mathcal{M}_{\text{search}} = \langle \{s^\star\}, \mathcal{A}_{\text{search}}, T_{\text{search}}, \mathcal{R}_{\text{search}}, \gamma \rangle
\] where $s^\star$ is a dummy world state and all nontrivial dynamics are pushed into the belief. An action $a_t \in \mathcal{A}_{\text{search}}$ corresponds to checking a location, which in practice bundles navigation and low-level manipulation into an atomic operation, e.g., \[a_t = \textsc{Check}(\ell_t), \] meaning go to $\ell_t$, open if necessary, and inspect it. The environment returns a textual observation $o_t$ (\textit{e.g.,} ``you see an apple'' or ``drawer is empty'').

\vspace{-10pt}
\paragraph{Belief-State Control Objective in Bayesian View}
Under the static-world assumption for the search phase, the true location of the target does not change; instead, only the agent's belief is updated. We model this by a belief update operator $\mathrm{BU}(\cdot)$ that defines the effective transition $T_{\text{search}}$ in belief space~\citep{kaelbling1998planning}, $b_{t+1} = \mathrm{BU}(b_t, a_t, o_t)$. The reward is sparse,
\[
\mathcal{R}_{\text{search}}(b_t, a_t, o_t)=
\begin{cases}
1, &\text{if target is found,}\\
0, &\text{otherwise,}
\end{cases}
\]
and the search task terminates either when the target is found or when a step budget is exhausted. Thus, each search task is a bandit-like, single-state MDP~\citep{rudra2022contextual} in which only the belief over locations evolves~\citep{kurenkov2023modeling}.

Within a full episode, multiple such search subtasks may appear (e.g., ``find the apple,'' then later ``find the mug''), each with its own candidate set $\mathcal{L}$ (possibly the same for different search subtasks) and initial belief $b_0$. The main ALFWorld/VirtualHome task, which may include navigation between rooms and object manipulation after the target is found, remains a POMDP, but our \emph{Align While Search} module operates specifically on the embedded search subtasks as an external belief module. Under this abstraction, our objective is to improve search efficiency: for a fixed environment and step budget, maximize the probability of successfully finding the target and minimize the expected number of search steps by explicitly maintaining and updating $b_t$.

From a Bayesian RL perspective~\citep{duff2002optimal}, this belief $b_t$ summarizes two coupled sources of epistemic uncertainty discussed in Section~\ref{sec:opportunities}: a latent environment type $\phi$ (e.g., tech-heavy vs. minimalist households) and the resulting object-location patterns conditioned on $\phi$. In principle, one could maintain a joint posterior $p(\phi, \ell \mid \tau_t)$ and plan in the augmented state space $(s^\star, b_t)$. In practice, AWS instantiates a structured approximation to this posterior: a global, language-level belief over latent environment hypotheses and a low-level belief over candidate check actions. The policy is then optimized in belief space, choosing actions based on their expected success and information gain with respect to these beliefs, which can be viewed as a contextual Bayesian bandit~\citep{agrawal2013thompson} adapted to single-state search subtasks.

\section{Proposed Method: Align While Search}
\label{sec:method}

\noindent Given the belief-augmented single-state search formulation in Section~\ref{sec:single_state_mdp}, we now describe \emph{Align While Search (AWS)}. AWS runs on top of an off-the-shelf LLM agent and is only invoked during search subtasks. It maintains a hierarchical belief over target-relevant structure, scores candidate \textsc{Check} actions using this belief, and returns the selected action to the base agent. We first introduce our belief parameterization, then explain how AWS performs exploratory action selection in belief space.

\subsection{Hierarchical Belief Representation}
\label{sec:belief_representation}

\noindent We represent the agent’s epistemic belief state with a hierarchical pair $(\mathcal{G}, \mathcal{S})$:
\begin{itemize}[itemsep=0pt, topsep=0pt, leftmargin=*]
    \item $\mathcal{G}$: a set of \textbf{global hypotheses} $B_t^\mathcal{G}$ over user habits and scene layout, stored as natural language (e.g., “the kitchen is well-organized; cabinets usually contain cups; mugs often appear near the sink”).
    \item $\mathcal{S}$: a \textbf{low-level belief} $b_t^\mathcal{S}$over candidate symbolic actions, stored as a categorical distribution over locations or receptacles (e.g., cabinet, countertop, drawer). An example distribution is visualized in Fig. \textcolor{blue}{6} in Appendix \textcolor{blue}{13.2}.
\end{itemize}

\noindent Concretely, the low-level belief $b^{\mathcal S}$ is modeled as
\begin{equation}
\label{eq:belief_s}
b_t^{\mathcal{S}}(a) = \Pr(\texttt{object found after } a),  \forall a \in \mathcal{L}^{\mathcal S}
\end{equation} 
where $\mathcal{L}^{\mathcal S}$ denotes the set of symbolic \textsc{Check} actions that inspect a particular location (bundling navigation and low-level manipulation into one step). The global hypotheses $\mathcal{G}$ are initialized once per episode from the initial observation using a single LLM prompt; the exact templates are given in Appendix \textcolor{blue}{13.2}

\vspace{-10pt}
\paragraph{Bayesian Latent-Variable View and Amortized Updates}
The belief introduced above implicitly summarizes two sources of epistemic uncertainty: a latent environment type $\phi$ (e.g., user/household cluster as in Figure~\ref{fig:heatmap_causal}) and the resulting object-location pattern. Let $z = (\phi, \ell)$ denote the joint latent variable, where $\ell$ indexes the target object location. In an ideal Bayesian formulation, we would maintain the posterior
\[
p(z \mid \tau_t) \propto p(o_t \mid a_t, z)\, p(z \mid \tau_{t-1}),
\]
and plan in the augmented state space $(s^\star, p(z \mid \tau_t))$, for example by maximizing information gain or expected task return. This latent-variable view is closely related to correlational object-search POMDPs~\citep{zheng2022towards}, which explicitly model a joint latent distribution over object locations and exploit correlational structure during planning.

In practice, AWS implements an \emph{amortized} approximation to this posterior~\citep{rezende2014stochastic}. Our hierarchical belief $(\mathcal{G}, \mathcal{S})$ induces a variational family
\[
q_\psi(z \mid \tau_t) = q_\psi(\phi \mid \tau_t)\, q_\psi(\ell \mid \phi, \tau_t),
\]
where $q_\psi(\phi \mid \tau_t)$ is represented by the textual global hypotheses $B^{\mathcal G}$ and $q_\psi(\ell \mid \phi, \tau_t)$ is represented by the action-level belief $b^{\mathcal S}$. The LLM-based update and projection operations,
\[
B^{\mathcal G}_{t} \xrightarrow{\pi_{BU}^g\, o_t} B^{\mathcal G}_{t+1}
\quad\text{and}\quad
B^{\mathcal G}_{t+1} \xrightarrow{\pi_{BP}^s} b^{\mathcal S}_{t+1},
\]
can be viewed as a black-box amortized inference map $\mathcal{F}_\psi(\cdot)$~\citep{kingma2013auto, kim2025qube}
\[
q_\psi(z \mid \tau_{t+1}) 
= \mathcal{F}_\psi\big(q_\psi(z \mid \tau_t), a_t, o_t\big)
\approx p(z \mid \tau_{t+1}),
\]
implemented by prompting a frozen LLM. AWS then performs control directly in belief space, selecting actions according to their expected task reward and information gain under $q_\psi$, which can be seen as a contextual Bayesian bandit defined over single-state search subtasks.

\subsection{Exploratory Action Selection}
\label{sec:exploratory_selection}

\noindent Given the current belief $b_t^{\mathcal S}$ (and associated global hypotheses $B^\mathcal{G}_t$), AWS scores each candidate action $a$ by its expected utility in belief space. We consider utilities that combine task reward and epistemic utility via information gain (IG). At time step $t$, we choose
\begin{equation}
\label{eq:astarsel}
a_t^* = \mathop{\arg\max}_{a \in \mathcal{A}_{\text{search}}} 
\mathbb{E}_{\hat{o} \sim p(\hat{o} \mid a, b_t)} 
\Big[ U\big(b_t, b_{t+1}(b_t, a, \hat{o})\big) \Big],
\end{equation}
where $b_{t+1}$ denotes the updated belief after taking $a$ and observing $\hat{o}$, and $U(\cdot)$ is a posterior utility. In this work, we instantiate $U$ as information gain,
\begin{equation}
\label{eq:ig}
\text{IG}(a) 
= \mathbb{E}_{\hat{o}} \big[ H(b_t) - H(b_{t+1} \mid a, \hat{o}) \big],
\end{equation}
where $H(\cdot)$ is the entropy of the action-level belief. The expectation over $\hat{o}$ is approximated using LLM-based observation simulation; details are given in Appendix \textcolor{blue}{13.1}

\begin{figure*}[t]
\vspace{-20pt}
    \centering
    \includegraphics[width=0.9\textwidth]{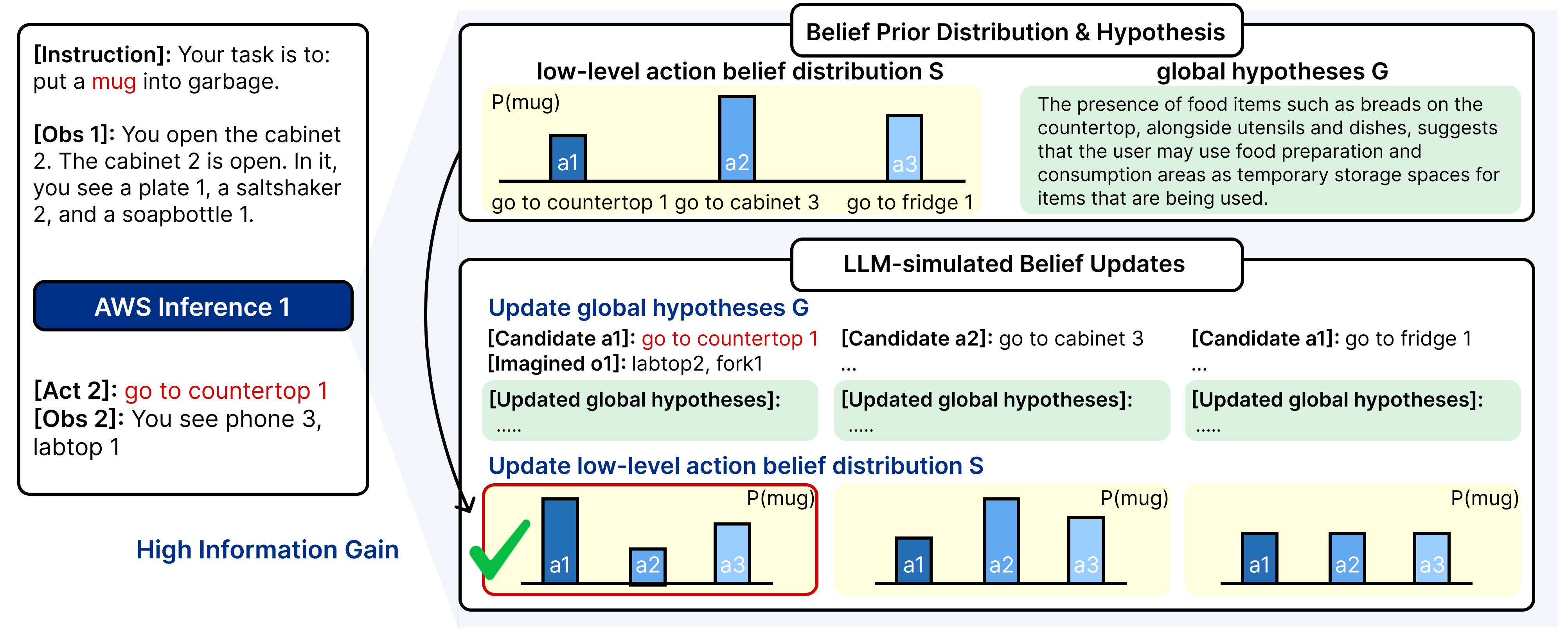}
    \caption{\textbf{Framework of Align While Search.}
    Given the current trajectory, AWS maps language to a numeric belief via update$\rightarrow$projection: observations update the global textual belief $B^{\mathcal G}$, which is then projected to a categorical action posterior $b^{\mathcal S}(a)$ (Eq.~\eqref{eq:belief_update}). Candidate actions are scored by expected IG (Eq.~\eqref{eq:ig}) under simulated observations $\hat{o}$, and the top-ranked action is returned to the base agent.}
    \label{fig:main_figure}
    \vspace{-20pt}
\end{figure*}

\vspace{-10pt}
\paragraph{LLM-simulated belief update.}
The belief state at time $t$ consists of textual global hypotheses $B_t^{\mathcal G}$ and numeric action-level scores $b_t^{\mathcal S}$. For a candidate action $a$ and a simulated observation $\hat{o}$, AWS updates belief in two stages:
\begin{equation}
\label{eq:belief_update}
B_t^{\mathcal G} \xrightarrow{\pi_{BU}^g(\hat{o})} B_{t+1}^{\mathcal G}
\quad\text{and}\quad
B_{t+1}^{\mathcal G} \xrightarrow{\pi_{BP}^s} b_{t+1}^{\mathcal S}.
\end{equation}
The first map $\pi_{BU}^g$ revises global hypotheses given $\hat{o}$ (e.g., removing “mugs are often in the sink” after observing an empty sink). The second map $\pi_{BP}^s$ projects the updated textual belief to an actionable distribution over symbolic locations. We consider two projection variants:
\begin{itemize}[itemsep=0pt, topsep=0pt, leftmargin=*]
    \item \textbf{Similarity-based projection:} adjusts belief scores using lexical similarity between candidate symbols and the updated hypotheses.
    \item \textbf{LLM-based projection:} queries an LLM to decide which symbols to boost or suppress given the hypotheses and location list.
\end{itemize}
Both variants induce different patterns of belief evolution (local refinement vs.\ semantic jumps), leading to distinct exploration behaviors; we analyze these qualitatively in Appendix \textcolor{blue}{21}. Prompt templates and exact score update rules (e.g., boost/suppress magnitudes) are provided in Appendix \textcolor{blue}{13}.

\subsection{Instance-Level Grounding and Termination}
\label{sec:termination}

\noindent The belief $b_t^{\mathcal S}$ is defined over symbolic locations (e.g., \texttt{cabinet}, \texttt{countertop}). After AWS selects a symbolic action, the underlying environment requires a concrete instance (e.g., \texttt{cabinet3}). We therefore sample an instance uniformly from the set of objects associated with the chosen symbol and execute the corresponding environment action. The environment then returns a reward and observation, and AWS performs an actual belief update (Eq.~\ref{eq:belief_update}) before proceeding to the next step.
\vspace{-12pt}
\paragraph{Termination.} The search loop terminates when the mission is satisfied (target object found and placed) or when a step budget is exhausted. In practice, we also use an observation-alignment score to detect when the agent has converged to a stable hypothesis; details of this criterion and the full pseudocode are provided in Appendix \textcolor{blue}{13.5}.

\begin{table*}[t]
  \centering
  \vspace{-10pt}
  \setlength{\tabcolsep}{4pt}
  \renewcommand{\arraystretch}{1.15}

  \begin{minipage}[t]{0.78\textwidth}\vspace{0pt}
    \raggedright
    \begin{tabular}{llccccccc}
    \toprule
    \multirow{2}{*}{\textbf{Backbone}} & \multirow{2}{*}{\textbf{Method}} &
    \multicolumn{6}{c}{\textbf{ALFWorld}} &
    \multirow{2}{*}{\textbf{Avg.}} \\
     & & CLEAN & COOL & EXAMINE & HEAT & PICK & PICK-2 & \\
    \midrule
    \multirow{6}{*}{GPT-4}
     & ReAct~\citep{yao2022react}         & 70.9 & 0.00 & 0.00 & 0.00 & 83.3 & 35.2 & 35.5 \\
     & Reflexion~\citep{shinn2023reflexion} & 64.5 & 90.5 & \textbf{100.} & \textbf{95.7} & 83.3 & 58.8 & 82.1 \\
     & RAP~\citep{hao2023reasoning}       & 70.9 & 9.52 & 5.55 & 4.34 & 79.1 & 29.4 & 33.2 \\ 
     & RAFA~\citep{liu2024reason}         & 35.5 & 47.6 & 55.6 & 43.5 & 58.3 & 5.90 & 41.0 \\
     & ReflAct~\citep{kim2025reflact}      & \textbf{96.8} & \underline{95.2} & \textbf{100.} & 78.3 & \textbf{95.8} & \textbf{94.1} & \textbf{93.3} \\
     
     & \cellcolor{blue!15}\textbf{AWS (Ours)}                & \cellcolor{blue!15}\textbf{96.8} & \cellcolor{blue!15}\textbf{100.0} & \cellcolor{blue!15}66.6 & \cellcolor{blue!15}\underline{91.3} & \cellcolor{blue!15}\underline{91.6} & \cellcolor{blue!15}\textbf{94.1} & \cellcolor{blue!15}\underline{90.0}\\
    \midrule
    \multirow{6}{*}{LLaMA-70B}
     & ReAct~\citep{yao2022react}         & 22.5  & 0.00  & 0.00 & 0.00 & 75.0 & 35.2 & 22.1 \\
     & Reflexion~\citep{shinn2023reflexion} & 54.8 & 61.9 & 27.8 & \textbf{78.3} & 50.0 & 35.3 & 51.3 \\
     & RAP~\citep{hao2023reasoning}       & 25.8 & 4.76 & 11.1 & 4.34 & 75.0 & 35.2 & 26.0 \\ 
     & RAFA~\citep{liu2024reason}         & \underline{61.3} & 38.1 & \underline{66.7} & 30.4 & 60.0 & 0.00 & 43.2 \\
     & ReflAct~\citep{kim2025reflact}      & 38.7 & \underline{66.7} & \textbf{72.2} & 56.5 & \underline{83.3} & \underline{52.9} & \underline{60.5} \\
     & \cellcolor{blue!15}\textbf{AWS (Ours)}                & \cellcolor{blue!15}\textbf{93.5} & \cellcolor{blue!15}\textbf{80.9} & \cellcolor{blue!15}44.4 & \cellcolor{blue!15}\underline{65.2} & \cellcolor{blue!15}\textbf{95.8} & \cellcolor{blue!15}\textbf{76.4} & \cellcolor{blue!15}\textbf{76.0}\\

    \bottomrule
    \end{tabular}
  \end{minipage}
  \vspace{-4pt}
  \hfill
  \begin{minipage}[t]{0.21\textwidth}\vspace{0pt}
    \centering
    \includegraphics[width=0.995\linewidth]{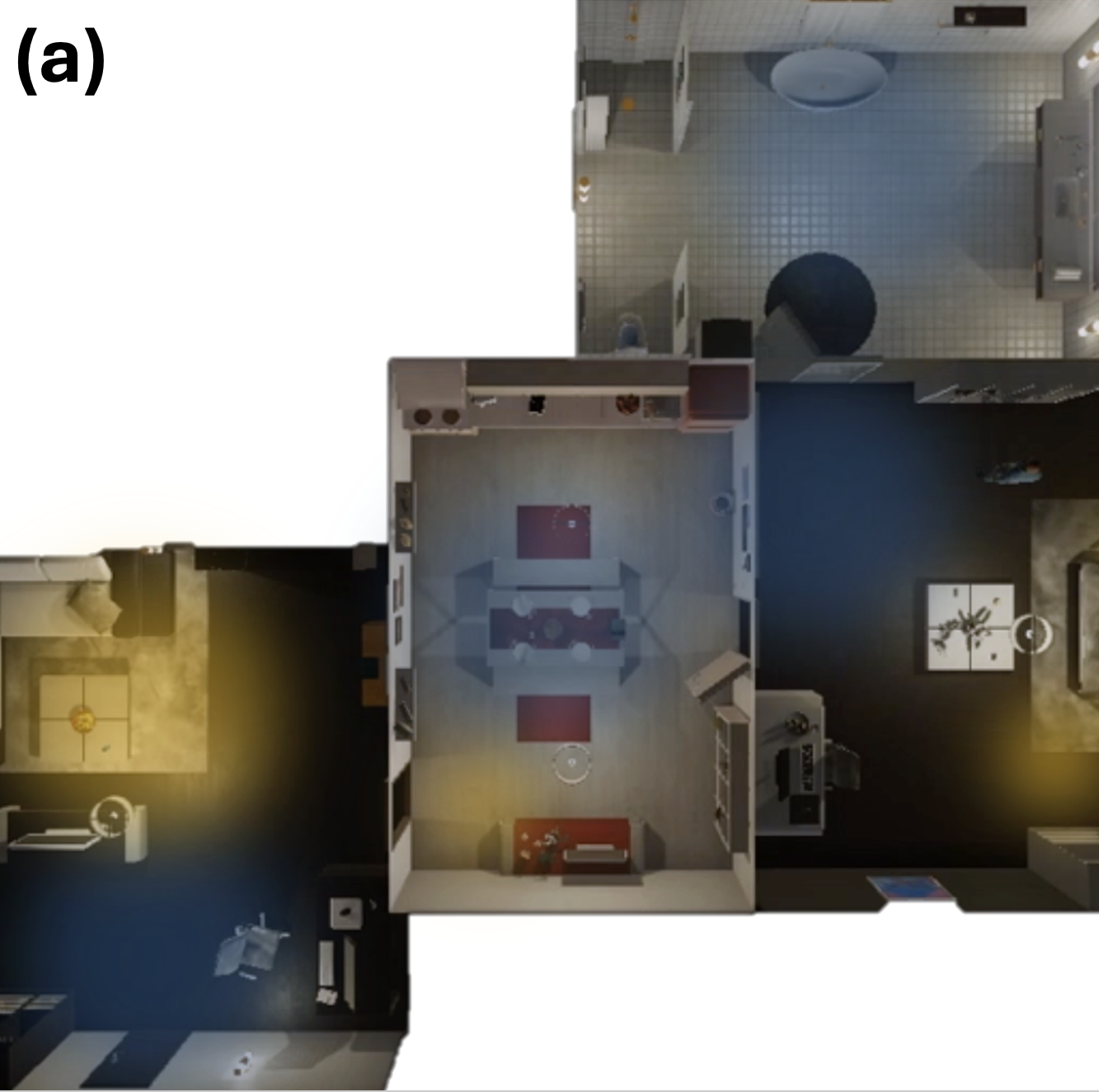}\\
    \includegraphics[width=0.995\linewidth]{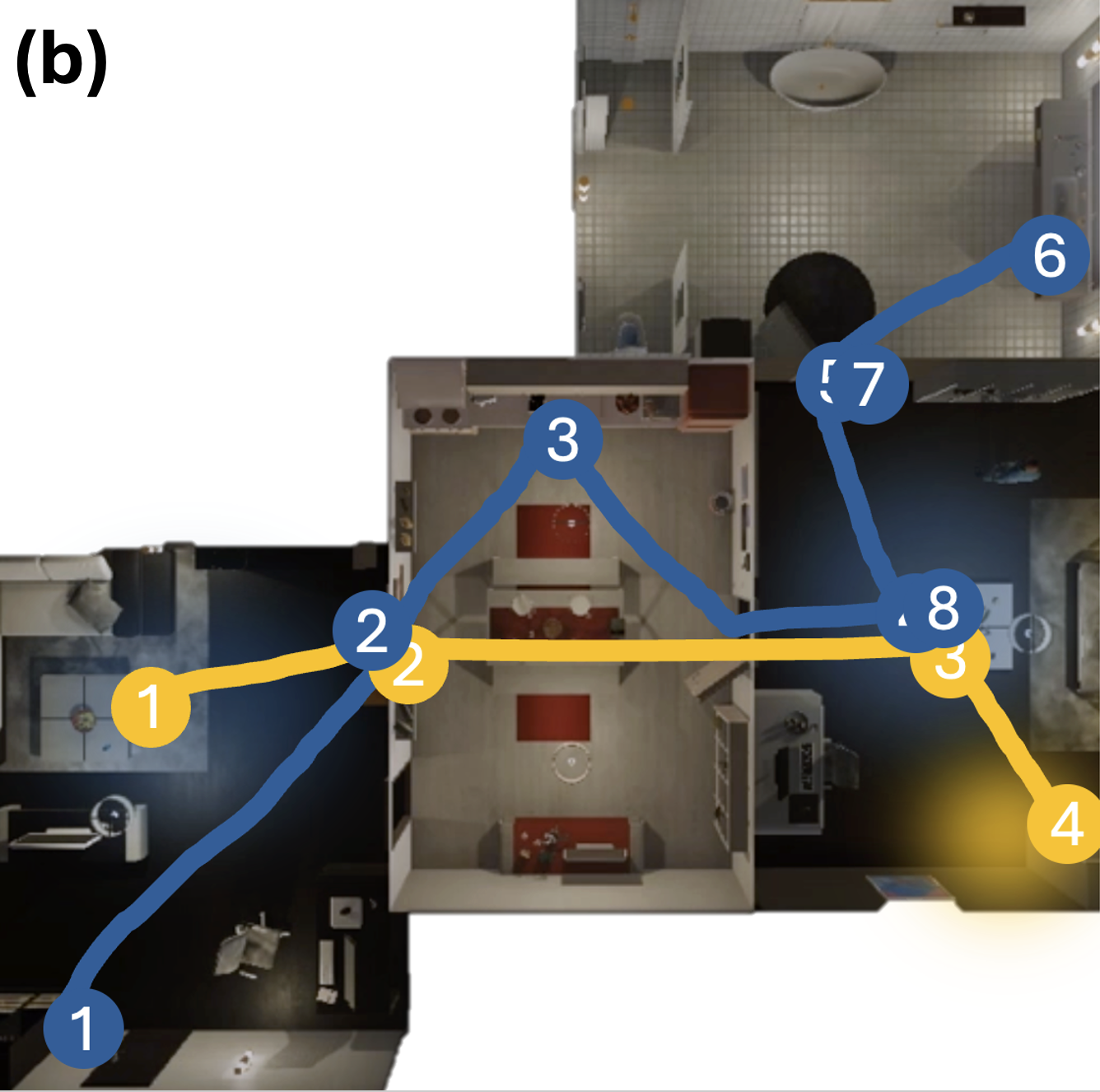}\\
  \end{minipage}
    \vspace{-4pt}
  \caption{(\textit{left}) ALFWorld sub-task performance compared to inference-time baselines across multiple LLM backbones (\textbf{Bold} indicates best performance, \underline{underline} indicates second-best); (\textit{right}) Illustration of belief-guided navigation in VirtualHome $(a)\to(b)$. ReAct (\textit{blue}) wanders through low-probability rooms, resulting in a long and inefficient trajectory (8 steps). AWS (\textit{yellow}) takes a substantially shorter route (4 steps) by committing early to a high-value hypothesis, corresponding to a rapid shift of belief mass from many low-probability rooms to a compact mode around the true target.}
  \vspace{-15pt}
  \label{tab:inference-detailed}
\end{table*}

\begin{table}[t]
\centering
\setlength{\tabcolsep}{1pt}
\begin{tabular}{lcc@{\hspace{12pt}}lcc}
\toprule
\multicolumn{3}{c}{\textbf{VirtualHome}} &
\multicolumn{3}{c}{\textbf{ALFWorld}} \\
\cmidrule(r){1-3}\cmidrule(l){4-6}
\textbf{Method} & \textbf{Seen} & \textbf{Unseen} &
\textbf{Method} & \textbf{Seen} & \textbf{Unseen} \\
\midrule
SFT~\citep{zeng2023agenttuning}        & 64.9 & 57.7 & SFT~\citep{zeng2023agenttuning}       & 79.3 & 71.6 \\
ETO\citep{song2024eto}   & 66.6 & 60.1 & ETO~\citep{song2024eto}       & 77.1 & 76.4 \\
IPR~\citep{xiong2024ipr}        & 67.6 & 61.9 & WKM~\citep{qiao2024agent}      & 77.1 & 78.2 \\
STeCa~\citep{wang2024steca}      & 69.6 & 63.6 & MPO~\citep{xiong2025mpo}      & 80.7 & 81.3 \\
\midrule
\textbf{AWS (Ours)}    & \textbf{69.6} & \textbf{65.2} &
\textbf{AWS (Ours)}    & \textbf{87.5} & \textbf{85.3} \\
\bottomrule
\end{tabular}
\caption{
\textbf{Comparison with train-time baselines on VirtualHome and ALFWorld environments.}
\textbf{Bold} indicates best performance.
}
\vspace{-20pt}
\label{tab:finetune-seen-unseen}
\end{table}

\section{Experiments}  

\noindent We evaluated the effectiveness of our belief-guided agent in simulated environments under partial observability. Our experiments were designed to answer three key questions:
\begin{itemize}
    \item Does inference-time belief refinement improve search efficiency compared to static or prior-only baselines?
    \item Is the predicted information gain (IG) aligned with actual improvements in belief accuracy?
    \item What factors affect the success or failure of belief-based action?
\end{itemize}

\subsection{Experimental Settings}
\paragraph{Environments} 
We evaluated our method on two benchmarks with sparse rewards and distinct types of action space: ALFWorld(text, image)~\citep{shridhar2020alfworld} and BabyAI~\citep{carta2023grounding} and VirtualHome(text, image)~\citep{puig2018virtualhome} . For high-level actions, we evaluated on \textbf{ALFWorld}~\citep{shridhar2020alfworld}, a household object search benchmark that requires goal-directed navigation and interaction. More generally, \textbf{VirtualHome}~\citep{puig2018virtualhome} provides a 3D, procedurally richer benchmark with unseen-scene generalization. We followed \citet{song2024eto} and \citet{wang2024steca} for trajectory construction and supervised training for each.  For low-level actions, we evaluated on \textbf{BabyAI}~\citep{carta2023grounding}, a Grid World environment in which agents and objects are placed in an 8x8 tile room (results in Appendix \textcolor{blue}{18.1}). 
Full dataset and environment details are provided in Appendix \textcolor{blue}{17}.
\vspace{-10pt}
\paragraph{Evaluation Metrics} 
We report the \textit{Success Rate}$(\%)$$(\uparrow)$ as our primary metric. We also report \textit{Average Steps} $(\downarrow)$ and Time/Token Cost $(\downarrow)$(cost analysis in Appendix \textcolor{blue}{20}) for comprehensive comparison.  
\vspace{-10pt}
\paragraph{Baselines} 
We compared our method (AWS) against three categories of baselines:  
(1) \textbf{Inference-time} baseline, such as ReAct~\citep{yao2022react}, Reflexion~\citep{shinn2023reflexion}, RAP~\citep{hao2023reasoning}, RAFA~\cite{liu2024reason}, and ReflAct~\cite{kim2025reflact};
(2) \textbf{Training-time} baselines, including ETO~\cite{song2024eto}, WKM~\cite{qiao2024agent}, and MPO~\cite{xiong2025mpo} for ALFWorld; IPR~\citep{xiong2024ipr} and STeCa~\citep{wang2024steca} for VirtualHome; (3) \textbf{LLM-based Embodied} baselines, including ZSP~\cite{huang2022language}, LLM-FT (a fine-tuned LLM), LLM-Planner~\cite{song2023llm}, SayCanPay~\cite{hazra2024saycanpay}, WorMI~\cite{yoo2025world} (results in Appendix \textcolor{blue}{18.2}). Verbal and schematic comparisons with AWS are provided in Appendix \textcolor{blue}{15}, and their reproduction details are presented in Appendix \textcolor{blue}{16}. 
\vspace{-13pt}
\paragraph{Model Backbones} 
 We evaluated AWS using both open-source and proprietary LLMs: LLaMA2-7B~\citep{grattafiori2024llama}, LLaMA3.1-8B~\citep{grattafiori2024llama},  LLaMA3.1-70B~\citep{grattafiori2024llama}, GPT-4o-mini~\citep{hurst2024gpt}, and GPT-4~\citep{hurst2024gpt}. Note that AWS requires $\pi_{pred}$, $\pi_{BU}^g$ and $\pi_{BP}^s$ and uses the same LLM for all three with different prompting. Implementation details, including the exact API variants used, decoding settings, and per-backbone performance variations, are provided in Appendix \textcolor{blue}{17}.

\subsection{Main Results}

\noindent As shown in Table~\ref{tab:inference-detailed}, Table~\ref{tab:finetune-seen-unseen} and Table \textcolor{blue}{7} (in Appendix \textcolor{blue}{18.1}), AWS provides strong inference-time gains over existing methods.
On ALFWorld (Table~\ref{tab:inference-detailed}), AWS achieves the highest average success rate with the LLaMA-70B backbone (76.0\%) and outperforms ReAct, RAP, RAFA, and Reflexion on almost all subtasks, while remaining highly competitive with GPT-4, where it ranks second in average performance (90.0\%) and attains the best or tied-best score on several subtasks (\textsc{Clean}, \textsc{Cool}, \textsc{Pick-2}).
Overall, AWS attains the best or tied-best performance on 7 out of 12 subtasks and stays within the top three on all subtasks.
On BabyAI, AWS also shows consistently strong results compared to inference-time methods (Table \textcolor{blue}{17}).

Compared to train-time baselines (Table~\ref{tab:finetune-seen-unseen}), AWS achieves the highest performance on both text-based VirtualHome and ALFWorld, reaching 69.6\%/65.2\% on VirtualHome and 87.5\%/85.3\% on ALFWorld, outperforming state-of-the-art methods such as STeCa and MPO, despite using only inference-time belief refinement without updating model weights.\footnote{We fine-tune LLaMA-2-7B and LLaMA-3.1-8B cas backbones for VirtualHome and ALFWorld, respectively.}
Hyperparameter sensitivity analysis is reported in Appendix \textcolor{blue}{18.3}.

\vspace{-13pt}
\paragraph{Generality on larger/multimodal settings} 

To assess whether our formulation extends beyond text-rendered environments, we also report AWS's search ability on the image-rendered multi-modal environment. In Table \textcolor{blue}{8} of Appendix \textcolor{blue}{18.2}, we further instantiate a multi-modal variant of AWS by pairing our belief module with a vision–language backbone. Without changing the search algorithm, this variant can operate on the same image-rendered environments, suggesting that our belief-guided exploratory inference is compatible with VLM-based embodied agents.

\subsection{Ablation Studies} \label{sec:ablation}

\begin{table}[htbp]
\centering
\renewcommand{\arraystretch}{0.95}
\setlength{\tabcolsep}{1pt}
\begin{tabular}{lcccc|cc}
\toprule
\textbf{Method} & \textbf{Prior} & \textbf{Update} & \textbf{IG} & \textbf{MCTS} & \textbf{SR (\%)} & \textbf{Steps ($\downarrow$)} \\
\midrule
Random Search          & $\times$ & $\times$ & $\times$ & $\times$ & 74.6 & 19.8 \\
Flat Prior Search      & \checkmark & $\times$ & $\times$ & $\times$ & 82.8 & 14.5 \\
Greedy (No IG)         & \checkmark & \checkmark & $\times$ & $\times$ & 82.4 & \textbf{11.7} \\
MCTS (No IG)           & \checkmark & \checkmark & $\times$ & \checkmark & \underline{85.0} & 14.7 \\
\textbf{Ours}          & \checkmark & \checkmark & \checkmark & \checkmark & \textbf{87.4} & \underline{13.8} \\
\bottomrule
\end{tabular}
\caption{\textbf{Ablations.} Removing either belief update or IG reduces alignment and success rate, confirming both are essential for inference-time performance (\textbf{bold} = best, \underline{underline} = second best).}
\label{tab:ablation}
\end{table}

\noindent To understand the contribution of individual components, we conducted controlled ablations and strategy replacements. Table~\ref{tab:ablation} reports the success rate and efficiency of alternative search strategies.
We tested four key variants:
(1) random unexplored action (random),
(2) no belief updates with greedy selection (fixed prior), 
(3) using a greedy action selection while updating beliefs (greedy), and
(4) using an MCTS-style action selection without IG estimation (MCTS). 

Our results show that belief updates are essential for maintaining alignment over time-removing them leads to significant drops in success and increases in episode length. Notably, belief updates alone do not yield gains (Flat Prior$\rightarrow$Greedy), but become evident when coupled with structured exploration. Similarly, removing IG (Ours$\rightarrow$MCTS) results in inefficient exploration, confirming that our lightweight IG predictor provides a strong signal. Together, these results highlight that posterior-guided exploration is most effective when belief updates, IG estimation, and structured search operate in concert, enabling accurate alignment inference-time behavior. 

\begin{table}[t]
\centering
\setlength{\tabcolsep}{6pt}
\begin{tabular}{llc}
\toprule
\textbf{Base Model} & \textbf{Method} & \textbf{ALFWorld} \\
\midrule
\multicolumn{3}{l}{\textbf{(A) Fine-tuned Models}} \\
\midrule
\multirow{3}{*}{\makecell[l]{LLaMA-3.1-8B\\\citep{zeng2023agenttuning}}} 
& \small Vanilla & 81.3 \\
& \quad \small LLM-Projection & 88.8~\textcolor{blue}{\scriptsize{(+9.20\%)}} \\
& \quad \small Similarity-Projection & \textbf{91.7}~\textcolor{blue}{\scriptsize{(+12.8\%)}} \\
\midrule
\multicolumn{3}{l}{\textbf{(B) Instruction-tuned Models}} \\
\midrule
\multirow{3}{*}{\makecell[l]{LLaMA-3.1-8B\\(base model)}} 
& \small Vanilla & 46.2 \\
& \quad \small LLM-Projection & 52.9~\textcolor{blue}{\scriptsize{(+14.5\%)}} \\
& \quad \small Similarity-Projection & \textbf{67.9}~\textcolor{blue}{\scriptsize{(+46.8\%)}}  \\
\midrule
\multirow{3}{*}{\makecell[l]{LLaMA-3.1-70B\\(base model)}} 
& \small Vanilla & 85.0 \\
& \quad \small LLM-Projection & 91.0~\textcolor{blue}{\scriptsize{(+7.06\%)}}  \\
& \quad \small Similarity-Projection & \textbf{94.0}~\textcolor{blue}{\scriptsize{(+10.6\%)}}  \\
\midrule
\multirow{3}{*}{\makecell[l]{GPT-4o-mini\\(base model)}} 
& \small Vanilla & 76.8 \\
& \quad \small LLM-Projection & 86.6~\textcolor{blue}{\scriptsize{(+12.7\%)}}  \\
& \quad \small Similarity-Projection & \textbf{89.5}~\textcolor{blue}{\scriptsize{(+16.5\%)}}  \\
\bottomrule
\end{tabular}
\caption{\textbf{Comparison of two projection methods.} Experiments across base models on ALFWorld-Text with MPO-inferred~\citep{xiong2025mpo}.}
\label{tab:main-result-b}
\end{table}

\vspace{-13pt}
\paragraph{Two Projection Variants.} Additionally, Table~\ref{tab:main-result-b} further summarizes the impact of our two projection strategies. Both LLM-based and similarity-based projections consistently boost performance across a wide range of models, including LLaMA-3.1-8B-SFT(+12.8\%), LLaMA-3.1-8B(+46.8\%), LLaMA-3.1-70B(+10.6\%), and GPT-4o-mini(+16.5\%), for both fine-tuned and instruction-tuned settings. Notably, our method synergizes with alternative world models such as MPO~\citep{xiong2025mpo}, achieving {\em new state-of-the-art results of 94.0\%} on text-based ALFWorld (Unseen). These results highlight that posterior-guided exploratory reasoning enables efficient inference-time adaptation and scalable deployment across environments and models.

\subsection{Computational Cost} \label{sec:computational_cost}  

\noindent Our method introduces higher inference-time overhead than ReAct due to LLM-based belief generation and updates. However, as in Figure \textcolor{blue}{7} in Appendix \textcolor{blue}{20}, it achieves comparable success rates$(\uparrow)$ with reduced average steps$(\downarrow)$, execution time$(\downarrow)$, and token usage per task$(\downarrow)$, against baselines, indicating more efficient trajectory generation.

\section{Discussion}

\subsection{How IG Changes Belief for Better Search}
\label{sec:ig_changes_belief}

\paragraph{Entropy dynamics across search.}
To understand how IG reshapes belief, we measure the entropy of location-level beliefs for both AWS and the SFT baseline, using the same LLaMA~3.1~8B SFT backbone. Let $\mathcal{L}$ be the set of location symbols (e.g., \texttt{cabinet}, \texttt{countertop}) and $\mathcal{A}_{\text{search}}$ the templated search actions. AWS maintains an explicit belief $b_t^{\mathcal S}$ over $\mathcal{L}$. For the SFT agent, which has no explicit belief state, we construct an \emph{implicit} belief by querying the backbone in scoring mode at each step $t$, obtaining log-probabilities $\log \pi_{\text{SFT}}(a\mid h_t)$ for all $a\in\mathcal{A}_{\text{search}}$, normalizing ($Z_t=\sum_{\ell'\in\mathcal{L}}\sum_{a:\,g(a)=\ell'}\pi_{\text{SFT}}(a\mid h_t)$) them with a single softmax, and aggregating probabilities over actions that inspect the same location $\ell$:
\[
b_t^{\text{SFT}}(\ell)
= \frac{1}{Z_t}\sum_{a:\,g(a)=\ell}\pi_{\text{SFT}}(a\mid h_t)
\]
For both agents we then compute the entropy
\[
H(b_t) = - \sum_{\ell\in\mathcal{L}} b_t(\ell)\log b_t(\ell),
\]
and define per-step information gain as $\Delta H_t = H(b_t)-H(b_{t+1})$, with net entropy reduction $\Delta H = H(b_0)-H(b_{\text{end}})$ between the beginning of search ($t=0$) and the step when the goal object is first found. We say that an episode exhibits \emph{overall belief sharpening} if the final belief is strictly sharper than the initial one ($H(b_{\text{end}})<H(b_0)$, and denote its frequency by $\Pr[\Delta H>0]$.
Table~\ref{tab:entropy_dynamics} summarizes three statistics averaged over successful episodes: the mean per-step information gain $\overline{\Delta H_t}$, the net entropy drop $\Delta H$, and the fraction of episodes with overall belief sharpening $\Pr[\Delta H>0]$. From a Bayesian perspective, the location-level belief $b_t$ can be viewed as an approximate posterior over the latent object location, so these quantities directly measure how effectively each agent reduces posterior uncertainty over the course of search. Empirically, AWS exhibits larger per-step updates and more than twice the cumulative entropy reduction of the SFT agent ($\Delta H$ 0.87 vs. 0.39), and belief sharpening occurs in a substantial higher fraction of episodes (84\% vs. 59\%). This indicates that IG drives more decisive and consistently beneficial reallocation of probability mass onto the correct locations during search. 

\vspace{-20pt}
\paragraph{Belief reshaping in a representative episode.}
Figure \textcolor{blue}{9} in Appendix \textcolor{blue}{21.1} illustrates this pattern on a \texttt{put mug in garbagecan} task. Panels (a)–(b) show belief over two symbols (\texttt{cabinet}, \texttt{countertop}) for similarity-based vs.\ LLM-based projection. The similarity-based agent updates beliefs smoothly and locally, while the LLM-based agent makes larger semantic jumps, briefly allocating probability to alternative receptacles. Panel (c) plots cabinet belief and entropy: for the LLM-based agent, entropy first rises when exploring semantically distant locations and then drops sharply once evidence supports cabinets, whereas the similarity-based agent shows milder changes. The corresponding action paths in panel (d) confirm this two-phase behavior: IG initially encourages broader exploration to de-bias wrong hypotheses, and subsequently focuses belief on a small set of plausible locations, enabling more efficient search.

\begin{table}[t]
\centering
\small
\setlength{\tabcolsep}{6pt}
\begin{tabular}{lccc}
\toprule
\textbf{Agent} 
& $\overline{\Delta H_t}$ 
& $\Delta H$ 
& $\Pr[\Delta H > 0]$ \\
\midrule
SFT (no IG) 
& 0.05 
& 0.39 
& 0.59 \\
AWS (ours) 
& \textbf{0.11} 
& \textbf{0.87} 
& \textbf{0.84} \\
\bottomrule
\end{tabular}
\caption{
\textbf{Entropy-based belief dynamics for the SFT baseline and AWS}. We use the same LLaMA~3.1~8B backbone. $\overline{\Delta H_t}$ denotes the average per-step information gain $\Delta H_t$.}
\vspace{-12pt}
\label{tab:entropy_dynamics}
\end{table}

\subsection{Further Discussion on Performance Gain}

\paragraph{Observation alignment increases throughout AWS.}
We measure an alignment reward that compares the LLM-predicted observation with the actual environment feedback after executing an action. As shown in Figure \textcolor{blue}{8}(left) in Appendix \textcolor{blue}{19}, this alignment score increases over search steps, indicating that AWS progressively refines its belief toward observations that are consistent with the true world state.
\vspace{-12pt}
\paragraph{Information gain and observation alignment correlates.}
We group steps into deciles according to their predicted IG score and track the change in alignment reward $\Delta R^{\text{align}}$. Higher-IG bins exhibit larger positive alignment gains (Figure \textcolor{blue}{8}(right) in Appendix \textcolor{blue}{19}), showing that IG serves as a reliable ordinal signal for belief-improving actions.

\begin{figure}[t]
\centering
\includegraphics[width=\linewidth]{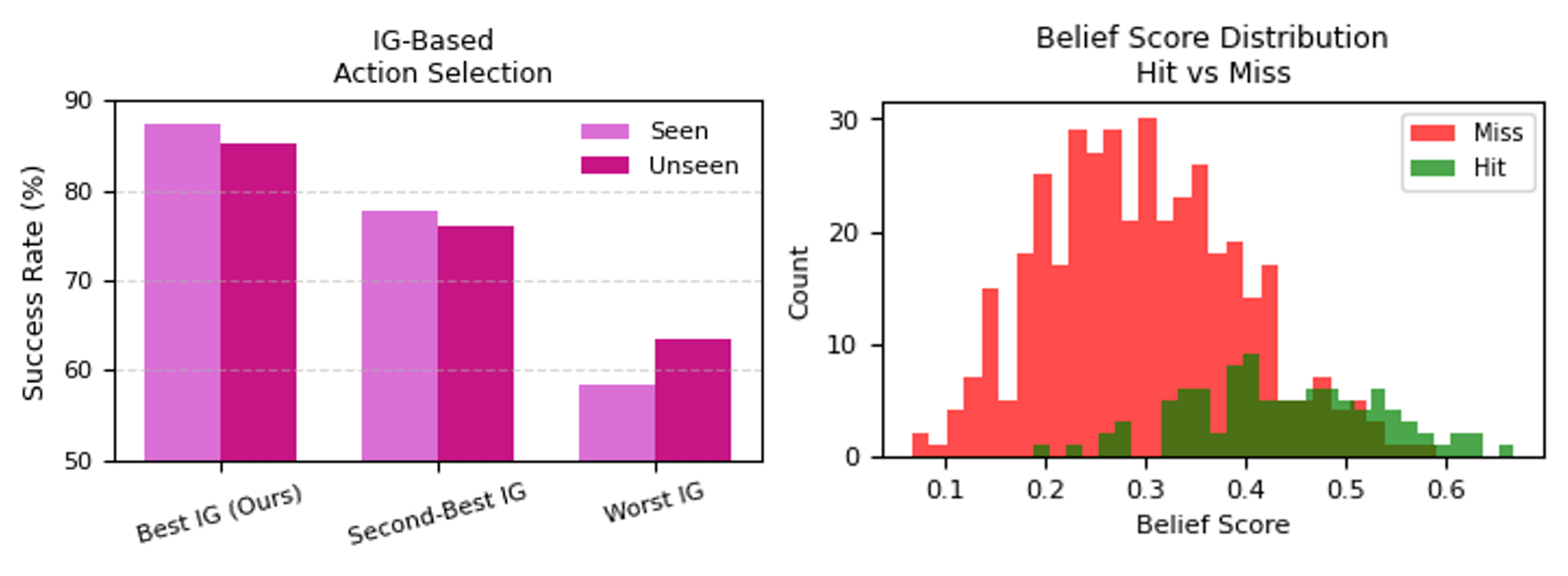}
\caption{
\textbf{Evaluating IG-Based Action Ranking and Belief Accuracy.} 
\textbf{(Left)} Success rate under different IG-based selection strategies. 
\textbf{(Right)} Belief scores at visited locations, grouped by whether the target object was present (hit) or absent (miss). 
}
\label{fig:ig_selection_ablation}
\vspace{-20pt}
\end{figure}
\vspace{-12pt}
\paragraph{Max information gain achieves the best in downstream task.}
To assess how IG affects control, we vary which action is chosen from the IG ranking. Selecting the top-IG action yields the highest success rates, while forcing the second-best or lowest-IG action leads to progressively worse performance (Figure \textcolor{blue}{5}(left); Appendix \textcolor{blue}{19}), confirming that IG provides an effective relative ranking for exploration.
\vspace{-12pt}
\paragraph{Belief prior aligns with actual object presence.}
We also compare belief scores at visited locations depending on whether the target object is present. Hit locations receive substantially higher probabilities than misses (Figure \textcolor{blue}{5}(right); $0.45$ vs.\ $0.31$, $p \ll 10^{-10}$; Appendix \textcolor{blue}{19}), indicating that the belief prior is well aligned with true object presence and actively guides the agent toward promising receptacles.

\subsection{Limitations}
\label{sec:limitations}

\noindent Our approach has several limitations. First, AWS inherits the strengths and weaknesses of the underlying language model. Performance depends on the base model's ability to simulate plausible household dynamics and object layouts; smaller models (e.g., Mistral-7B, DeepSeek-8B) often fail to generate reliable hypothetical observations, and their IG rankings collapse toward nearly static priors. In addition, AWS increases test-time cost: for each search step, we simulate observations and perform belief updates for multiple candidate actions, which adds wall-clock and token overhead compared to a single-pass SFT agent.

Second, our formulation explicitly targets a single-state search regime where the physical world is effectively static during search and episodes factor into \textsc{find-then-act} subtasks. This matches ALFWorld and VirtualHome, but does not directly handle environments with non-stationary object locations, concurrent agents, or more entangled long-horizon objectives.

Finally, the belief update and information-gain computation rely on hand-designed surrogates. The approximate posterior $q_\psi(z \mid \tau)$ is implemented via prompting and similarity-based or boost/suppress projection rules, and IG is used primarily as an ordinal ranking signal rather than a calibrated quantity. As a result, AWS offers no formal optimality guarantees, and its behavior can be sensitive to prompt design and hyperparameters; learning world models and trained belief updaters is an important direction for future work.

\section{Conclusion}
This work proposes a belief-augmented single-state search module that uses information gain in belief space to guide LLM agents under partial observability. By maintaining hierarchical hypotheses over environment structure and updating them at inference time, AWS turns static prompt-based agents into adaptive explorers, improving success and search efficiency in embodied benchmarks. Future work includes learning the belief updater and extending AWS beyond static search subtasks to dynamic, multimodal environments and richer forms of world alignment.

\clearpage 
\newpage

{
    \small
    \bibliographystyle{ieeenat_fullname}
    \bibliography{main}
}

\clearpage 

\end{document}